\documentclass[11pt]{article}

\usepackage[english]{babel}
\usepackage[utf8]{inputenc}
\usepackage[
	pdftitle={Efficient Hyperparameter Tuning for Large Scale Kernel Ridge Regression},
	pdfauthor={Giacomo Meanti},
	breaklinks=true
]{hyperref}
\usepackage{breakcites}
\usepackage{graphicx}
\usepackage{svg}
\usepackage[inline]{enumitem}

\usepackage{amssymb}
\usepackage{amsmath}
\usepackage{amsthm}
\usepackage{mathtools}
\usepackage{bm}
\usepackage{suffix}
\usepackage{empheq}
\usepackage{xcolor}
\usepackage{xfrac}
\usepackage{booktabs}
\usepackage{multirow}
\usepackage{authblk}
\usepackage[a4paper,
			left=1in, right=1in, top=1in, bottom=1in,
			bindingoffset=0.2in,
			footskip=.25in]{geometry}

\usepackage{etoolbox,siunitx}
\robustify\bfseries
\usepackage[
	sorting=nyt,
	backend=biber,
	style=authoryear-comp,
	hyperref=true,
	natbib=true,
	maxcitenames=2,
	maxbibnames=99,
	giveninits=true,
	uniquename=init,
	parentracker=true,
	url=false,
	doi=false,
	eprint=true,
]
{biblatex}
\graphicspath{{figures/}}

\usepackage{amssymb}
\usepackage{amsmath}
\usepackage{amsthm}
\usepackage{mathtools}
\usepackage{bm}
\usepackage{suffix}
\usepackage{empheq}
\usepackage{xcolor}

\newcommand{\nystrom}{Nystr{\"o}m}

\newcommand{\hilbert}{\ensuremath{\mathcal{H}}}

\definecolor{malgared}{rgb}{0.8627450980392157, 0.34901960784313724, 0.34901960784313724}
\definecolor{malgagray}{rgb}{0.2275, 0.2275, 0.2275}

\DeclarePairedDelimiter\autoVbar{\lVert}{\rVert}

\DeclarePairedDelimiter\autobracket{(}{)}
\DeclarePairedDelimiter\autosqbracket{[}{]}
\DeclarePairedDelimiter\autovbar{\lvert}{\rvert}

\newcommand{\norm}[1]{\ensuremath{\autoVbar{#1}}}
\WithSuffix\newcommand{\norm}*[1]{\ensuremath{\autoVbar*{#1}}}

\DeclareMathOperator{\Ima}{Im}

\newcommand{\eye}{\ensuremath{\bm{I}}}

\newcommand{\real}[1]{\mathbb{R}^{#1}}
\newcommand{\defeq}{\vcentcolon=}  
\DeclarePairedDelimiterX{\infdivx}[2]{[}{]}{%
	#1\;\delimsize|\delimsize|\;#2%
}

\newcommand{\tr}[1]{\ensuremath{\mathrm{Tr}\autobracket*{#1}}}
\WithSuffix\newcommand\tr*[1]{\ensuremath{\mathrm{Tr}\autobracket{#1}}}

\renewcommand{\det}[1]{\ensuremath{\autovbar*{#1}}}

\DeclareMathOperator*{\argmin}{arg\,min}

\newcommand{\expect}[1]{\ensuremath{\mathbb{E}\autosqbracket*{#1}}}

\makeatletter
\newcommand{\pushright}[1]{\ifmeasuring@#1\else\omit$\displaystyle#1$\ignorespaces\fi}
\newcommand{\pushleft}[1]{\ifmeasuring@#1\else\omit$\displaystyle#1$\hfill\fi\ignorespaces}
\makeatother

\newcommand{\valloss}[1]{\ensuremath{\mathcal{L}_V\autobracket{#1}}}

\WithSuffix\newcommand{\valloss}*[1]{\ensuremath{\mathcal{L}_V^*\autobracket{#1}}}
\newcommand{\hp}{\ensuremath{\bm{\lambda}}}
\WithSuffix\newcommand{\hp}*{\ensuremath{\bm{\lambda}^*}}

\newcommand{\la}{\ensuremath{\lambda}}

\newcommand{\kappr}{\ensuremath{\widetilde{K}}}

\newcommand{\knm}{\ensuremath{K_{nm}}}

\newcommand{\kmm}{\ensuremath{K_{mm}}}

\newcommand*\numcircledmod[1]{\raisebox{.5pt}{\textcircled{\raisebox{-.9pt} {#1}}}}

\newcommand{\nysfh}{\ensuremath{\hat{f}_{\la, Z, \gamma}}}

\newcommand{\fh}{\ensuremath{\hat{f}_{\la, \gamma}}}
\newcommand{\f}{\ensuremath{f_{\la, \gamma}}}
\newcommand{\kf}{k_\gamma}
\newcommand{\errhat}[1]{\hat{L}(#1)}
\newcommand{\err}[1]{L(#1)}
\newcommand{\errhatl}[1]{\hat{L}_{\la}(#1)}
\renewcommand{\hm}{\hilbert_m}
\newcommand{\spa}[1]{\mathrm{span}\{#1\}}
\newcommand{\reg}[1]{\la\norm{#1}_\hilbert^2}

\newcommand{\nkrr}{\mbox{N-KRR}}

\newtheorem*{theorem*}{Theorem}
\newtheorem{lemma}{Lemma}
\newtheorem*{lemma*}{Lemma}

\newtheorem{remark}{Remark}

\title{Efficient Hyperparameter Tuning for Large Scale Kernel Ridge Regression}
\author[1]{Giacomo Meanti}
\author[1]{Luigi Carratino}
\author[2]{Ernesto De Vito}
\author[1,3,4]{Lorenzo Rosasco}
\affil[1]{\small \textit{MaLGa, DIBRIS, Università degli Studi di Genova}}
\affil[2]{\small \textit{MaLGa, DIMA, Università degli Studi di Genova}}
\affil[3]{\small \textit{CBMM, Massachusetts Institute of Technology}}
\affil[4]{\small \textit{Istituto Italiano di Tecnologia}}

\begin{document}
	\maketitle
	\begin{refsection}
		\vspace*{-1em}
	\begin{abstract}
  		Kernel methods provide a principled approach to nonparametric learning. 
		While their basic implementations scale poorly to large problems, recent advances showed that approximate solvers can efficiently handle 
		massive datasets.
		A shortcoming of these  solutions is that hyperparameter tuning is not taken care of, and left for the user to perform.
		Hyperparameters are crucial in practice and the lack of automated tuning greatly hinders efficiency and usability. 
		In this paper, we work to fill in this gap focusing on kernel ridge regression based on the Nystr\"om approximation.
		After reviewing and contrasting a number of  hyperparameter tuning strategies,
		we propose a        complexity regularization criterion based on a data dependent penalty, and discuss its efficient optimization.
		Then, we proceed to a careful and extensive empirical evaluation highlighting strengths and weaknesses of the different tuning strategies. 
		Our analysis shows the benefit of the proposed approach, that we hence incorporate in a library for large scale kernel methods to derive adaptively tuned solutions.
	\end{abstract}

\section{Introduction}

Learning from finite data requires fitting models of varying complexity to training data. 
The problem of finding the model with the right complexity is referred to as model selection in statistics and more broadly  as hyperparameter tuning in machine learning.
The problem is classical and known to be of utmost importance for machine learning algorithms
to  perform well in practice. The literature in statistics is extensive \citep{esl}, including a number of theoretical results \citep{tsyabakov03, arlotphd, massart07}. Hyperparameter (HP) tuning is also at the core of recent trends such as 
neural architecture search \citep{elsken19nas} or AutoML \citep{automl_book}.
In this paper, we consider the question of hyperparameter tuning in the context of kernel methods 
and specifically kernel ridge regression (KRR) \citep{smola00}. Recent advances showed that kernel methods can be scaled to massive data-sets using approximate solvers \citep{eigenpro2, hierachical17, billions}. The latter take advantage of a number of ideas from optimization \citep{boyd04} and randomized algorithms \citep{alaoui14}, and exploit parallel computations with GPUs. While these solutions open up new possibilities for applying kernel methods, hyperparameter tuning is notably missing, ultimately hindering their practical use and efficiency. 
Indeed, available solutions which provide hyperparameter tuning are either limited to small data, or are restricted to very few hyperparameters \citep{liquisvm17, scikit-learn,lssvm02}.

In this paper we work to fill in this gap. We consider approximate solvers based on the Nystr\"om approximation and work towards an automated tuning of the regularization and kernel parameters, as well as the \nystrom{} centers. On the one hand, we provide a careful review and extensive empirical comparison for a number of hyperparameter tuning strategies, while discussing their basic theoretical guarantees. 
On the other hand we propose, and provide an efficient implementation for, a novel criterion inspired by complexity regularization \citep{bartlett02} and based on a data-dependent bound. 
This bound treats separately the sources of variance due to the stochastic nature of the data. In practice, this results in better stability properties of the corresponding tuning strategy. As a byproduct of our analysis we complement an existing library for large-scale kernel methods with the possibility to adaptively tune a large number of hyperparameters. Code is available at the following address: \url{https://github.com/falkonml/falkon}.

In Section~\ref{sec:background} we introduce the basic ideas behind empirical risk minimization and KRR, as well as hyperparameter tuning. In Section~\ref{sec:new-obj} we propose our new criterion, and discuss its efficient implementation in Section~\ref{sec:approximations}. In Section~\ref{sec:experiments} we conduct a thorough experimental study and finally, in Section~\ref{sec:conclusion} we provide some concluding remarks.

	\section{Background}\label{sec:background}
	
We introduce the problem of learning a model's parameters, which leads to learning of the hyperparameters and then discuss various objective functions and optimization algorithms which have been proposed for the task. 

\subsection{Parameter and Hyperparameter Learning}\label{sec:hp-tuning}
Assume we are given a set of measurements $\{(x_i, y_i)\}_{i=1}^n \subset \mathcal{X}\times\mathcal{Y}$ related to each other by an unknown function $f^*: \mathcal{X}\rightarrow\mathcal{Y}$ and corrupted by some random noise $\epsilon_i$ with variance $\sigma^2$.
\begin{equation}\label{eq:fixed-des}
	y_i = f^*(x_i) + \epsilon_i.
\end{equation}
We wish to approximate the target function $f^*$ using a model $f:\mathcal{X}\rightarrow\mathcal{Y}$ defined by a set of \textit{parameters} which must be learned from the limited measurements at our disposal. In order for the learning procedure to succeed, one often assumes that $f$ belongs to some hypothesis space $\mathcal{F}$, and this space typically depends on additional \textit{hyperparameters} $\theta$.
Assume we are given a loss function $\ell: \mathcal{Y}\times\mathcal{Y}\rightarrow\mathbb{R}$; we can learn a model by fixing the hyperparameters $\theta$ and minimizing the loss over the available training samples:
\begin{equation*}
	\hat{f}_\theta = \argmin_{f\in\mathcal{F}_\theta} \sum_{i=1}^n \ell(f(x_i), y_i)
\end{equation*}
In this paper we are concerned with kernel ridge regression: a specific kind of model where the loss function is the squared loss $\ell(y, y') = \norm{y - y'}^2$ and the hypothesis space is a reproducing kernel Hilbert space (RKHS) $\hilbert$. Associated to $\hilbert$ is a kernel function $\kf: \mathcal{X}\times\mathcal{X}\rightarrow\real{}$ which depends on hyperparameters $\gamma$. To ensure that the minimization problem is well defined we must add a regularization term controlled by another hyperparameter $\lambda$:
\begin{equation*}
	\fh = \argmin_{f\in\hilbert} \sum_{i=1}^n \norm{f(x_i) - y_i}^2 + \reg{f}.
\end{equation*}
The solution to this minimization problem is unique~\citep{caponnetto07}, but is very expensive to compute requiring $O(n^3)$ operations and $O(n^2)$ space.
An approximation to KRR considers a lower-dimensional subspace $\hm\subset\hilbert$ as hypothesis space, where $\hm$ is defined from $m\ll n$ points $Z=\{z_j\}_{j=1}^m \subset \mathcal{X}$ \citep{williams01}. 
While the inducing points $Z$ (also known as \nystrom{} centers) are often picked from the training set with different sampling schemes~\citep{kumar12}, they can also be considered as hyperparameters. In fact this is common in sparse Gaussian Processes (GPs) and leads to models with a much smaller number of inducing points~\citep{titsias09, hensman13, hensman15}.
Minimizing the regularized error gives the unique solution 
\begin{equation}\label{eq:n-krr-beta}
	\nysfh = \sum_{j=1}^m \beta_j \kf(\cdot, z_j),\quad \text{with}~~\beta = (\knm^\top\knm + \lambda n \kmm)^{-1}\knm^\top Y
\end{equation}
with $(\knm)_{i,j} = \kf(x_i, z_j)$ and $(\kmm)_{i,j} = \kf(z_i, z_j)$. 
The \nystrom{} KRR model (N-KRR) reduces the computational cost of finding the coefficients to $O(n\sqrt{n}\log n)$ when using efficient solvers~\citep{rudi2017falkon,billions,eigenpro2}.

The ideal goal of hyperparameter optimization is to find a set of hyperparameters $\theta^*$ for which $\hat{f}_{\theta^*}$ minimizes the test error (over all unseen samples). 
By definition we cannot actually evaluate the test error: we must use the available data points. 
Na{\"i}vely one could think of minimizing the training error instead, but such a scheme inevitably chooses overly complex models which overfit the training set. 
Instead it is necessary to minimize a data-dependent criterion $\mathcal{L}$
\begin{equation*}
	\widehat{\theta} = \argmin_\theta \mathcal{L}(\hat{f}_\theta)
\end{equation*}
such that complex models are penalized. In practice a common strategy for choosing $\mathcal{L}$ is for its expectation (with respect to the sampling of the data) to be equal to, or an upper bound of the test error.
In the next section we will look at instances of $\mathcal{L}$ which appear in the literature and can be readily applied to N-KRR.

\subsection{Objective Functions}\label{sec:objectives}

\paragraph{Validation error}
A common procedure for HP tuning is to split the available $n$ training samples into two parts: a training set and a validation set. The first is used to learn a model $\hat{f}_\theta$ with fixed hyperparameters $\theta$, while the validation set is used to estimate the performance of different HP configurations.
\begin{equation}
\mathcal{L}^{\mathrm{Val}}(\hat{f}_\theta) = \frac{1}{n_\mathrm{val}}\sum_{i=1}^{n_\mathrm{val}} \norm{\hat{f}_\theta(x_i^\mathrm{val}) - y_i^\mathrm{val}}^2
\end{equation}
By using independent datasets for model training and HP selection, $\mathcal{L}^{\mathrm{Val}}$ will be an unbiased estimator of the test error and it can be proven that its minimizer is close to  $\theta^*$ under certain assumptions~\citep{arlot09}. However, since $\hat{f}_\theta$ has been trained with $n_\mathrm{tr} < n$ samples, there is a small bias in the chosen hyperparameters~\citep{varma06}. Furthermore the variance of the hold-out estimator is typically very high as it depends on a specific data split.
Two popular alternatives which address this latter point are k-fold cross-validation (CV) which averages over $k$ hold-out estimates and leave-one-out CV.

\paragraph{Leave-one-out CV and Generalized CV}
The LOOCV estimator is an average of the $n$ estimators trained on all $n-1$ sized subsets of the training set and evaluated on the left out sample.
The result is an almost unbiased estimate of the expected risk on the full dataset~\citep{vapnikslt}.  
For linear models a computational shortcut allows to compute the LOOCV estimator by training a single model on the whole dataset instead of $n$ different ones~\citep{cawley2004}. In particular in the case of \nkrr{} we can consider
\begin{equation}\label{eq:nkrr-loocv}
\mathcal{L}^{\mathrm{LOOCV}}(\hat{f}_\theta) = \frac{1}{n}\sum_{i=1}^n\Bigg( \dfrac{y_i - \hat{f}_\theta(x_i)}{1 - H_{ii}} \Bigg)^2,
\end{equation}
where the so-called hat matrix $H$ is $H = \knm(\knm^\top\knm + \lambda n \kmm)^{-1}\knm$.

GCV is an approach proposed in \citet{golub79} to further improve LOOCV's computational efficiency and to make it invariant to data rotations:
\begin{equation}\label{eq:nkrr-gcv}
\mathcal{L}^{\mathrm{GCV}}(\hat{f}_\theta) = \frac{1}{n}\sum_{i=1}^n \Bigg( \dfrac{y_i - \hat{f}_\theta(x_i)}{ \frac{1}{n} \tr{I - H} }\Bigg)^2.
\end{equation}
For GCV \citet{cao06} proved an oracle inequality which guarantees convergence to the neighborhood of $\theta^*$ when estimating $\lambda$ for KRR.

\paragraph{Complexity regularization}
Complexity regularization, or covariance penalties~\citep{mallow73, efron04covariance} are a general framework for expressing objective functions as the empirical error plus a penalty term to avoid overly complex models.
For linear models the trace of the hat matrix acts as penalty against complexity. 
Applying these principles to N-KRR gives the objective
\begin{equation}\label{eq:creg-obj}
\mathcal{L}^{\mathrm{C-Reg}}(\nysfh) = \frac{1}{n}\norm{\nysfh(X) - Y}^2 + \dfrac{2\sigma^2}{n}\tr{(\kappr + n\la I)^{-1}\kappr}
\end{equation}
where $\kappr = \knm\kmm^\dagger\knm^\top$ (the \nystrom{} kernel), and $A^\dagger$ denotes the Moore-Penrose inverse of matrix $A$. The first term can be interpreted as a proxy to the bias of the error, and the second as a variance estimate.
For estimating $\la$ in (N-)KRR, \citet{arlotbach09} proved an oracle inequality if a precise estimate of the noise $\sigma^2$ is available.

\paragraph{Sparse GP Regression \citep{titsias09}}
A different approach comes from a Bayesian perspective, where the equivalent of KRR is Gaussian Process Regression (GPR). Instead of estimating the test error, HP configurations are scored based on the ``probability of a model given the data'' \citep{gpbook}. 
A fully Bayesian treatment of the hyperparameters allows to write down their posterior distribution, from which the HP likelihood has the same form of the marginal likelihood in the model parameter's posterior. 
Hence maximizing the (log) marginal likelihood (MLL) with gradient-based methods is common practice in GPR.

Like with N-KRR, inducing points are used in GPR to reduce the computational cost, giving rise to models such as SoR, DTC, FiTC~\citep{candela05}. Here we consider the SGPR model proposed in \citet{titsias09} which treats the inducing points as variational parameters, and optimizes them along with the other HPs by maximizing a lower bound to the MLL. The objective to be minimized is 
\begin{equation}\label{eq:sgpr}
\mathcal{L}^{\mathrm{SGPR}}(\nysfh) = \log\det{\kappr + n\lambda\eye}
+ Y^\top(\kappr + n\lambda\eye)^{-1}Y + \frac{1}{n\lambda} \tr{K - \kappr}.
\end{equation}
The first term of Eq.~\eqref{eq:sgpr} penalizes complex models, the second pushes towards fitting the training set well and the last term measures how well the inducing points approximate the full training set. 
Recently the approximate MLL was shown to converge to its exact counterpart~\citep{burtconv20}, but we note that this does not guarantee convergence to the optimal hyperparameters.

\subsection{Optimization Algorithms}\label{sec:algos}
In this section we describe three general approaches for the optimization of the objectives introduced above.

\paragraph{Grid search}
In settings with few hyperparameters the most widely used optimization algorithm is grid-search which tries all possible combinations from a predefined set, choosing the one with the lowest objective value at the end. 
Random search~\citep{bergstra11} and adaptive grid search (used for SVMs in \citet{liquisvm17}) improve on this basic idea, but they also become prohibitively costly with more than $\sim 5$ HPs as the number of combinations to be tested grows exponentially.

\paragraph{Black-box optimization}
A more sophisticated way to approach the problem is to take advantage of any smoothness in the objective. Sequential model-based optimization (SMBO) algorithms~\citep{brochu10, snoek12, shahriari16} take evaluations of the objective function as input, and fit a Bayesian \textit{surrogate} model to such values. The surrogate can then be cheaply evaluated on the whole HP space to suggest the most promising HP values to explore.
These algorithms do not rely on gradient information so they don't require the objective to be differentiable and can be applied for optimization of discrete HPs.
However, while more scalable than grid search, black-box algorithms become very inefficient in high (i.e. $> 100$) dimensions.

\paragraph{Gradient-based methods}
Scaling up to even larger hyperparameter spaces requires exploiting the objective's local curvature. While the optimization problem is typically non-convex, gradient descent will usually reach a good local minimum. When the objective can be decomposed as a sum over the data-points SGD can be used, which may provide computational benefits (e.g. the SVGP objective~\citep{hensman13} is optimized in mini-batches with SGD).
In the context of KRR, gradient-based methods have been successfully used for HP optimization with different objective functions~\citep{seeger08, keerthi07svmho}.
Recent extensions to gradient-based methods have been proposed for those cases when the trained model cannot be written in closed form. Either by unrolling the iterative optimization algorithm~\citep{maclaurin15,franceschi17,grazzi20}, or by taking the model at convergence with the help of the implicit function theorem~\citep{pedregosa16,rajeswaran19}, it is then possible to differentiate a simple objective (typically a hold-out error) through the implicitly defined trained model.
This has proven to be especially useful for deep neural nets~\citep{lorraine20}, but is unnecessary for N-KRR where the trained model can be easily written in closed form.

	\section{Hyper-parameter Optimization for \nystrom{} KRR}\label{sec:new-obj}
	\begin{figure}
	\centering
	\includesvg[width=.65\linewidth]{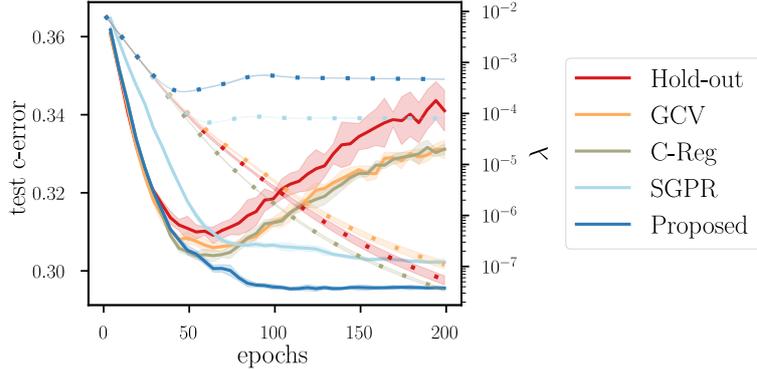}
	\caption{Test-error and penalty ($\la$) as a function of optimization epoch on the small-HIGGS dataset. $m=100$ centers, $d$ lengthscales and $\la$ were optimized with equal initial conditions.  The three unbiased proxy functions lead to overfitting, while SGPR and the proposed objective do not.}
	\label{fig:optim-higgs}
\end{figure}

The objectives introduced in the previous section can be applied to HP tuning for kernel methods.
Always keeping in mind efficiency but also usability, our goal is to come up with an objective and associated optimization algorithm which:
\begin{enumerate*}[label=\arabic*)]
	\item can be used to tune the hyperparameters of \nystrom{} kernel ridge regression including the inducing points and
	\item can be computed efficiently, even for large scale problems.
\end{enumerate*}

To satisfy the first point, an algorithm of the first-order is needed since the inducing points are typically between a hundred and a few thousands (each point being of the same dimension as the data). 
Regarding the second point we found empirically that the unbiased objectives are prone to overfitting on certain datasets. An example of this behavior is shown in Figure~\ref{fig:optim-higgs} on a small subset of the HIGGS dataset.
The first three objectives (Hold-out, GCV and C-Reg) are unbiased estimates of the test error, hence it is their variance which causes overfitting.
To mitigate such possibility in our objective we may look into the different sources of variance: \textit{hold-out} depends strongly on which part of the training set is picked for validation, \textit{GCV} and \textit{C-Reg} don't rely on data splitting but still suffer from the variance due to the random initial choice of inducing points.

We set out to devise a new objective in the spirit of complexity regularization, which is an upper bound on the test error.
A biased estimate -- which is therefore overpenalizing -- will be more resistant to noise than an unbiased one (as was noted in \citet{arlotphd}), and we tailor our objective specifically to N-KRR in order to explicitly take into account the variance from inducing point selection.

We base our analysis of the N-KRR error in the fixed design setting, where the points $x_i \in \mathcal{X}, i=(1, \dots, n)$ are assumed to be fixed, and the stochasticity comes from i.i.d. random variables $\epsilon_i, \dots, \epsilon_n$ such that $\expect{\epsilon_i} = 0$ and $\expect{\epsilon_i^\top\epsilon_i} = \sigma^2$.
Denote the empirical error of an estimator $f\in\hilbert$ as $\errhat{f} = n^{-1}\norm{f(X) - Y}^2$ and the test error as $\err{f} = n^{-1}\norm{f(X) - f^*(X)}^2$ (recall $f^*$ from Eq.~\eqref{eq:fixed-des}).
Consider inducing points $z_j$ and a subspace of $\hilbert$: $\hm = \spa{\kf(z_1, \cdot), \dots, \kf(z_m, \cdot)}$, $m\ll n$, and let $P$ be the projection operator with range $\hm$.
Denote the regularized empirical risk as $\errhatl{f} = \errhat{f} + \reg{f}$, 

Assessing a particular hyperparameter configuration $(\la, Z, \gamma)$ requires estimating the expected test error at the empirical risk minimizer trained with that configuration $\nysfh$; the optimal HPs then are found by $(\la, Z, \gamma)^* = \argmin_{(\la, Z, \gamma)} \err{\nysfh}$. The following lemma gives an upper bound on the ideal objective; a full proof is available in Appendix~\ref{app:derivation}.
\begin{lemma}\label{lem:main}
	Under the assumptions of fixed-design regression we have that,
	\begin{align}
		\expect{\err{\nysfh}} \le &\nonumber \dfrac{2\sigma^2}{n}\tr{(\kappr + n\la I)^{-1} \kappr} \nonumber \\
		& + \dfrac{2}{n\la}\tr{K - \kappr} \expect{\errhat{\f}} \nonumber \\
		& + 2\expect{\errhat{\f}} \label{eq:ub}
	\end{align}
\end{lemma}
\textit{Proof sketch.~~}We decompose the test error expectation in the following manner
\begin{align*}
	& \expect{\err{\nysfh}} \le \mathbb{E}\Big[\underbrace{\err{\nysfh} - \errhat{\nysfh}}_{\numcircledmod{1}} \\
	& + \underbrace{\errhat{\nysfh} + \reg{\nysfh} - \errhatl{P\f}}_{\numcircledmod{2}} 
	+ \underbrace{\errhatl{P\f}}_{\numcircledmod{3}} \Big]
\end{align*}
by adding and subtracting $\errhat{\nysfh}$, $\errhatl{P\f}$ and summing the positive quantity $\la\norm{\nysfh}_{\hilbert}^2$.
Since $\nysfh$ is the minimizer of $\errhat{\nysfh} + \reg{\nysfh}$ in the space $\hm$ and since $P\f\in\hm$, the second term is negative and can be discarded.

Term \numcircledmod{1} is the variance of \nkrr{} and can be computed exactly by noting that
\begin{align*}
	\expect{\errhat{\nysfh}} &= \expect{n^{-1}\norm{\nysfh(X) - f^*(X) - \epsilon}^2} \\
	&= \expect{\err{\nysfh}} + \sigma^2  -\frac{2}{n}\expect{\langle \nysfh(X) - f^*(X), \epsilon \rangle}
\end{align*}
where the first part cancels and we can ignore $\sigma^2$ which is fixed and positive. Expanding the inner product and taking its expectation we are left with
\begin{align*}
\frac{2}{n}\expect{\langle \nysfh(X) - f^*(X), \epsilon \rangle} = \frac{2\sigma^2}{n}\tr{(\kappr + n\la I)^{-1}\kappr}
\end{align*}
which is the \textit{effective dimension} or the \textit{degrees of freedom} of the hypothesis space $\hm$, times the noise variance $\sigma^2$.

Term \numcircledmod{3} takes into account the difference between estimators in $\hilbert$ and in $\hm$. 
We begin by upper-bounding the regularized empirical error of $P\f$ with a first part containing the projection operator and a second term without $P$
\begin{equation*}
	\expect{\errhat{P\f} + \reg{P\f}}
	 \le \expect{\frac{2}{n}\norm{K^{1/2}(I-P)}^2\norm{\f}^2 + 2\errhatl{\f}}.
\end{equation*}
Now $\norm{K^{1/2}(I - P)}^2 \le \tr*{K - \kappr}$ the difference between full and approximate kernels, and $\norm{\f}^2 \le \la^{-1} \errhatl{\f}$ which leads us to the desired upper bound.\qed

We now make two remarks on computing Eq.~\eqref{eq:ub}.
\begin{remark}{(Computing $\expect{\errhatl{\f}}$)}
\textnormal{In the spirit of complexity regularization we can approximate this bias term by the empirical risk of \nkrr{} $\errhatl{\nysfh}$, so that the final objective will consist of a data-fit term plus two complexity terms: the effective dimension and the \nystrom{} approximation error.}
\end{remark}

\begin{remark}{(Estimating $\sigma^2$)}
\textnormal{Once again following the principle of overpenalizing rather than risking to overfit, we note that in binary classification the variance of $Y$ is capped at 1 for numerical reasons, while for regression we can preprocess the data dividing $Y$ by its standard deviation. Then according to Eq.~\eqref{eq:fixed-des} we must have that the label standard deviation is greater than the noise standard deviation hence $\hat{\sigma}^2 = 1 \ge \sigma^2$.}
\end{remark}

Our final objective then has a form which we can compute efficiently
\begin{align}
	 \mathcal{L}^{\mathrm{Prop}} = &\dfrac{2}{n}\tr{(\kappr + n\la I)^{-1} \kappr} \nonumber \\
	& + \dfrac{2}{n\la}\tr{K - \kappr} \errhatl{\nysfh} \nonumber \\
	& + \dfrac{2}{n}\norm{\nysfh(X) - Y}^2 + \lambda\norm{\nysfh}_{\hilbert}^2. \label{eq:fin-obj}
\end{align}
We make two further remarks on the connections to the objectives of Section~\ref{sec:objectives}.
\begin{remark}{(Similarities with complexity regularization)}
	\textnormal{
		$\mathcal{L}^\mathrm{Prop}$ has a similar form to Eq.~\eqref{eq:creg-obj} with an extra term which corresponds to the variance introduced by the \nystrom{} centers which we were aiming for (up to multiplication by the KRR bias).
	}
\end{remark}

\begin{remark}{(Similarities with SGPR)}
	\textnormal{
		Eq.~\eqref{eq:fin-obj} shares many similarities with the SGPR objective: the log-determinant is replaced by the model's effective dimension -- another measure of model complexity -- and the term $\tr*{K-\kappr}$ is present in both objectives. Furthermore the data-fit term in $\mathcal{L}^{\mathrm{SGPR}}$ is
		\begin{align*}
		Y^\top (\kappr + n\la I)^{-1} Y &= \dfrac{1}{\la} (n^{-1}\norm{\nysfh(X) - Y}^2 + \reg{\nysfh}) \\
		& = \dfrac{1}{\la} \errhatl{\nysfh}
		\end{align*}
		which is the same as in the proposed objective up to a factor $\la^{-1}$.
	}
\end{remark}

	\section{Scalable Approximations}\label{sec:approximations}
\begin{figure}
	\centering
	\includesvg[width=.7\linewidth]{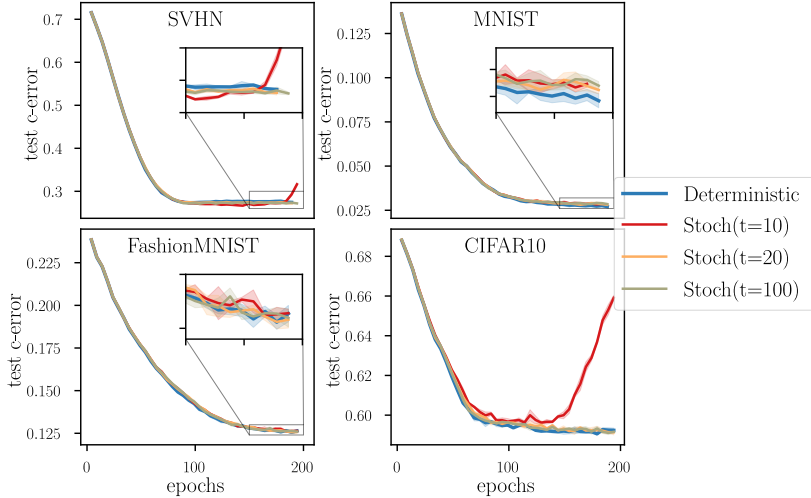}
	\caption{The effect of stochastic trace estimation. We plot the optimization curves of the exact objective $\mathcal{L}^\mathrm{Prop}$ (\textit{Deterministic}) and the approximated objectives with 10, 20 and 100 STE vectors. On the four datasets we optimized $m=200$ centers, $\la$ and $\gamma$.}
	\label{fig:ste}
\end{figure}

	Some practical considerations are needed to apply the objective of Eq.~\eqref{eq:fin-obj} to large-scale datasets -- for which direct computation is not possible due to space or time constraints.
We examine the terms comprising $\mathcal{L}^{\mathrm{Prop}}$ and discuss their efficient computations.
In Figure~\ref{fig:ste}, we verify that the resulting approximation is close to the exact objective.

Starting with the last part of the optimization objective (the one which measures data-fit) we have that
\begin{equation*}
	\norm{\nysfh(X)-Y}^2 + \la\norm{\nysfh}_{\hilbert}^2
		= Y^\top (I - \underbrace{\knm (\overbrace{\knm^\top\knm+n\la\kmm}^{B})^{-1}\knm^\top) Y}_{=\nysfh(X)}
\end{equation*}
which can be computed quickly using a fast, memory-efficient N-KRR solver such as Falkon~\citep{billions} or EigenPro~\citep{eigenpro2}. 
However we must also compute the objective's gradients with respect to all HPs, and since efficient solvers proceed by iterative minimization, such gradients cannot be trivially computed using automatic differentiation, indeed, it would be in principle possible to unroll the optimization loops and differentiate through them, the memory requirements for this operation would be too high for  large datasets.

\paragraph{Efficient gradients}
A solution to compute the gradients efficiently is to apply the chain rule by hand until they can be expressed in terms of matrix vector products $(\nabla K) \bm{v}$ with $K$ any kernel matrix (i.e.~$\knm$ or $\kmm$) and $\bm{v}$ a vector.
As an example the gradient of the data-fit term is
\begin{equation*}
	\nabla (Y^\top \knm B^{-1}\knm^\top Y) = 2 Y^\top (\nabla \knm)B^{-1}\knm^\top Y - Y^\top \knm B^{-1} (\nabla B) B^{-1} \knm^\top Y
\end{equation*}
where we can obtain all $B^{-1}\knm^\top Y$ vectors via a non-differentiable \nkrr{} solver, and multiply them by the (differentiable) kernel matrices for which gradients are required.
Computing these elementary operations is efficient, with essentially the same cost as the forward pass $K\bm{v}$, and can be done row-wise over $K$.
Block-wise computations are essential for low memory usage since kernel matrices tend to be huge but kernel-vector products are small, and they allow trivial parallelization across compute units (CPU cores or GPUs).
In many cases these operations can also be accelerated using KeOps~\citep{keops}.

The remaining two terms of Eq.~\eqref{eq:fin-obj} are harder to compute. 
Note that in $\tr*{K - \kappr}$ we can often ignore $\tr*{K}$ since common kernel functions are trivial when computed between a point and itself, but more in general it only requires evaluating the kernel function $n$ times. We thus focus on
\begin{equation}\label{eq:comp-obj-tr}
	\tr{\kappr} = \tr{\knm\kmm^{\dagger}\knm^\top}
\end{equation}
and on the effective dimension
\begin{equation}\label{eq:comp-obj-deff}
	\tr{(\kappr + \la I)^{-1}\kappr} = \tr{\knm B^{-1}\knm^\top}.
\end{equation}
Both these terms are traces of huge $n\times n$ matrices. By their symmetry we can express them as squared norms reducing the space requirements to $n\times m$, but they still remain slow to compute: just the $\knm^\top\knm$ term costs more than training a \nkrr{} model with the Falkon solver.

\paragraph{Trace estimation}
A simple approximation can vastly improve the efficiency of computing Equations \eqref{eq:comp-obj-tr}, \eqref{eq:comp-obj-deff}, and  their gradients: stochastic trace estimation (STE). 
The Hutchinson estimator~\citep{hutchinson} approximates $\tr*{A}$ by $\frac{1}{t}\sum_{i=1}^t r_i^\top A r_i$ where $r_i$ are zero mean, unit standard deviation random vectors.
We can use this to estimate Eq.~\eqref{eq:comp-obj-deff} by running the Falkon solver with $R = [r_1, \dots, r_t]$ instead of the labels $Y$ to obtain $(\knm^\top\knm + \la\kmm)^{-1}\knm^\top R$, then multiplying the result by $\knm^\top R$ and normalizing by the number of stochastic estimators $t$. 
The same random vectors $R$ can be used to compute $\knm^\top R$ for Eq.~\eqref{eq:comp-obj-tr}, coupled with the Cholesky decomposition of $\kmm$. STE reduces the cost for both terms from $O(nm^2)$ to $O(nmt)$ which is advantageous since $t<m$.
In Figure~\ref{fig:all_stdy} we investigate whether the approximate objective matches the exact one, and how $t$ affects the approximation.
The observed behavior is that as few as 10 vectors are enough to approximate the full objective for a large part of the optimization run, but it can happen that such coarse approximation causes the loss to diverge. Increasing $t$ to 20 solves the numerical issues, and on all the datasets tested we found $t=20$ to be sufficient.

Alternatively, Eq.~\eqref{eq:comp-obj-tr} can be approximated with a \nystrom{}-like procedure: taking a random subsample of size $p$ from the whole dataset, denote $K_{pm}$ as the kernel matrix between such $p$ points and the $m$ \nystrom{} centers; then 
$$\tr{\knm\kmm^\dagger\knm^\top} \approx \frac{n}{p} \tr{K_{pm}\kmm^\dagger K_{pm}^\top}$$
which can be computed in $pm^2 + m^3$ operations. By choosing $p\sim m$ the runtime is then $O(m^3)$, which does not depend on the dataset size, and is more efficient than the STE approach. 
Unfortunately, this additional \nystrom{} step cannot be effectively applied for computing Eq.~\eqref{eq:comp-obj-deff} where the inversion of $B$ is the most time-consuming step.

	\section{Experiments}\label{sec:experiments}

To validate the objective we are proposing for HP optimization of N-KRR models we ran a series of experiments aimed at answering the following questions:
\begin{enumerate}
	\item Since our objective is an upper-bound on the test error, is the over-penalization acceptable, and what are its biases?
	\item What is its behavior during gradient-based optimization: does it tend to overfit, does it lead to accurate models?
	\item Does the approximation of Section~\ref{sec:approximations} enable us to actually tune the hyperparameters on large datasets?
\end{enumerate}

\begin{figure}
	\centering
	\includesvg[width=0.6\textwidth]{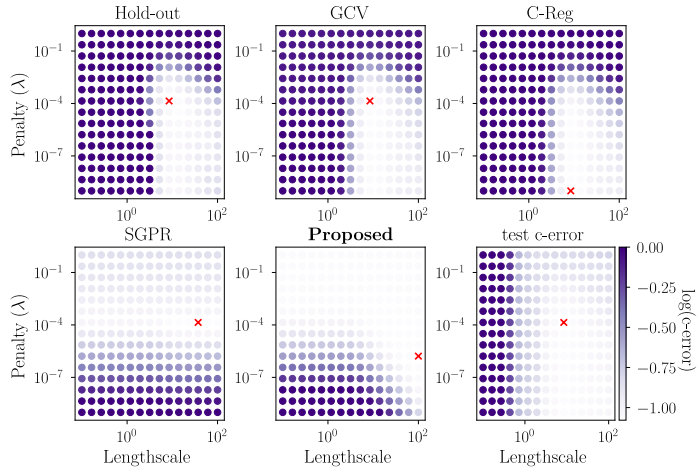}
	\caption{Effectiveness of test error proxies on a grid. The objective values (log transformed) are plotted at different $\lambda, \gamma$ points for the \textit{small-HIGGS} dataset. Lighter points indicate a smaller objective and hence a better hyperparameter configuration. The minimum of each objective is denoted by a cross.
	}
\label{fig:grid_search}
\end{figure}

The first point is a sanity check: would the objective be a good proxy for the test error in a grid-search scenario over two hyperparameters ($\lambda$ and $\gamma$ with the RBF kernel). 
This doesn't necessarily transfer to larger HP spaces, but gives an indication of its qualitative behavior. 
In Figure~\ref{fig:grid_search} we compare 5 objective functions to the test error on such 2D grid. 
It is clear that the three functions which are unbiased estimators of the test error have very similar landscapes. 
Both SGPR and the proposed objective instead have the tendency to \textit{overpenalize}: SGPR strongly disfavors low values of $\lambda$, while our objective prefers high $\lambda$ and $\gamma$. 
This latter feature is associated with simpler models: a high $\gamma$ produces smooth functions and a large $\lambda$ restricts the size of the hypothesis.

We will see that the subdivision of objective functions into two distinct groups persists during optimization. However, in general it will not be true that the unbiased objectives produce models with lower test error than the overpenalized ones. The best performing method is going to depend on the dataset.

\paragraph{Small-scale optimization}
\begin{figure*}[ht!]
	\centering
	\includesvg[width=0.97\textwidth]{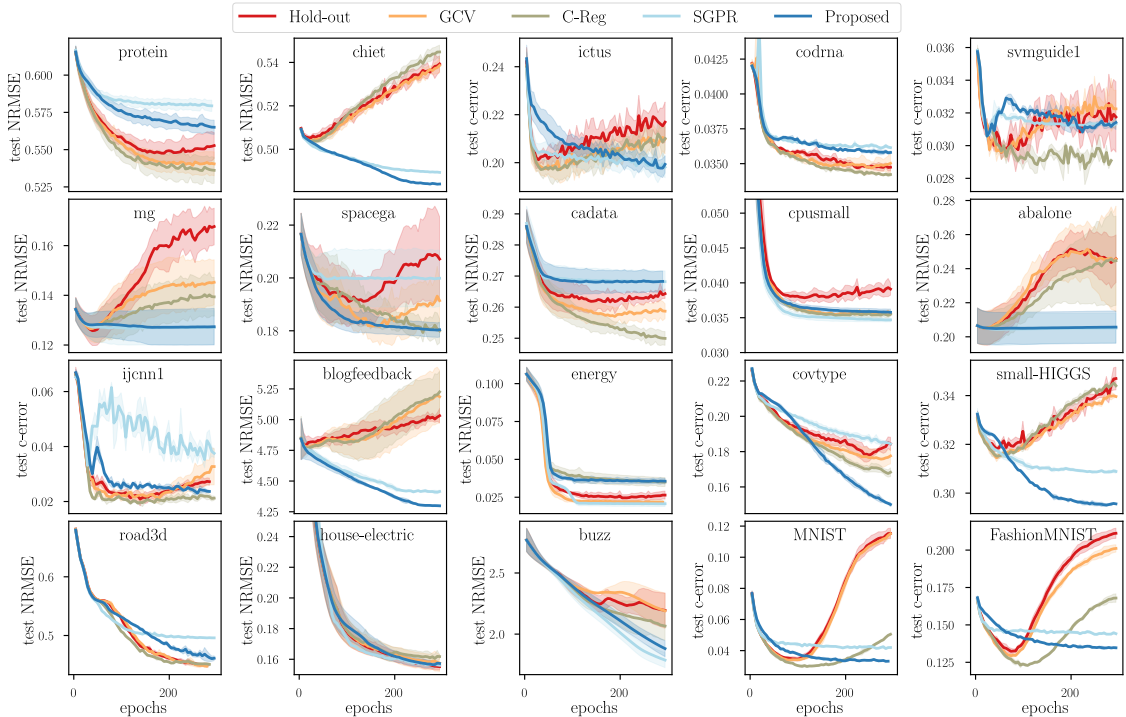}
	\caption{Empirical comparison of five objective functions for hyperparameter tuning. On each dataset we optimized $m=100$ \nystrom{} centers, a separate lengthscale for each dimension and $\la$ for 200 epochs with a learning rate of 0.05 using the Adam optimizer. Also reported is the standard deviation from 5 runs of the same experiment with a different random seed. Each dataset has its own error metric. Labels of regression datasets were normalized to have unit standard deviation. }
	\label{fig:all_stdy}
\end{figure*}
We used the exact formulas, along with automatic differentiation and the Adam optimizer to minimize the objectives on 20 datasets taken from the UCI repository, the LibSVM datasets, or in-house sources (more details on the datasets in Appendix~\ref{app:datasets}).
We automated the optimization runs as much as possible to avoid having to set many meta-hyperparameters: fixed learning rate, the initial value for $\la$ set to $1/n$ and the initial value for $\gamma$ set with the median heuristic~\citep{medianheuristic}. 
We used early stopping when the objective values started increasing. 
The results -- shown in Figure~\ref{fig:all_stdy} -- confirm our previous observations: there are some datasets (among which \textit{small-HIGGS}, \textit{buzz}, \textit{house-electric}) on which the unbiased objectives overfit the training set while the proposed proxy function does not. In fact in some cases the hyperparameters found with our objective are much better than the ones found, for example, with the C-Reg objective.
On the other hand, there is another group of datasets (e.g.~\textit{protein}, \textit{energy} or \textit{codrna}) where the extra bias of the proposed objective becomes detrimental as the optimization gets stuck into a suboptimal configuration with higher test error than what would be attainable with an unbiased objective.

Among the three unbiased objectives, hold-out clearly performs the worst. This is due to its high variance, and could be mitigated (at the expense of a higher computational cost) by using k-fold cross-validation. 
The GCV and C-Reg objectives perform similarly to each other in many cases. Especially in the image datasets however, GCV overfits more than C-Reg. 

SGPR closely matches the proposed objective as it doesn't overfit. However, on several datasets it produces worse HPs than our objective displaying a larger bias. 
On the other hand there are other datasets for which the ranking is reversed, so there is no one clear winner. We must note however that the SGPR objective cannot be efficiently computed due to the log-determinant term, when datasets are large.

\paragraph{Large-scale optimization}
We tested the performance of the proposed objective with STE on three large-scale datasets, comparing it against two variational sparse GP solvers~\citep{gpflow17,gpytorch18} which also learn a compact model with optimized inducing points and a classic \nkrr{} model with lots of randomly chosen centers trained with Falkon. 
Our tests are all performed in comparable conditions, details available in Appendix~\ref{app:experimental}.
The results in Table~\ref{tab:large-scale} tell us that we can approach (but not quite reach) the performance -- both in terms of speed and accuracy -- of a very large model using a small fraction of the inducing points. 
They also support the conclusion that our objective is effective at optimizing a large number of hyperparameters, at least on par with methods in the GPR framework. 
\begin{table}[t]
	\caption{Error and running time of kernel solvers on large-scale datasets. We compare our objective with two approximate GPR implementations and hand-tuned \nkrr{} (Falkon).} \label{tab:large-scale}
	\centering
		\begin{tabular}{p{3.5em} l c c c c}
			\toprule
			& & $\mathcal{L}^{\mathrm{Prop}}$ & GPyTorch & GPFlow & Falkon\\
			\cmidrule{3-6}
			
			\multirow{3}{3.5em}{Flights $n\approx10^6$}     & error    & 0.794 & 0.803 & 0.790 & 0.758  \\
			& time(s)  & 355   & 1862  & 1720  & 245    \\
			& m        & 5000  & 1000  & 2000  & $10^5$ \\
			\cmidrule{2-6}
			\multirow{3}{3.5em}{Flights-Cls $n\approx10^6$} & error    & 32.2  & 33.0  & 32.6  & 31.5   \\
			& time(s)  & 310   & 1451  & 627   & 186    \\
			& m        & 5000  & 1000  & 2000  & $10^5$ \\
			\cmidrule{2-6}
			\multirow{3}{3.5em}{Higgs $n\approx10^7$}       & error    & 0.191 & 0.199 & 0.196 & 0.180  \\
			& time(s)  & 1244  & 3171  & 1457  & 443    \\
			& m        & 5000  & 1000  & 2000  & $10^5$ \\
			\bottomrule
		\end{tabular}
\end{table}

	\section{Conclusions}\label{sec:conclusion}
	In this paper, we improved the usability of large scale kernel methods proposing a gradient-based solution for tuning a large number of hyperparameters, on large problems. We incorporate this method into an existing library for large scale kernel methods with GPUs.
	We showed that it is possible to train compact \nystrom{} KRR models if the centers are allowed to deviate from the training set, which can speed up inference by orders of magnitude.
	A future work will be to consider complex parametrized kernels which allow to improve the state of the art of kernel-based models on structured datasets such as those containing images or text.
	
	\section*{Acknowledgments}
	The authors would like to thank the Anonymous Reviewers for their helpful comments on trace approximation. Lorenzo Rosasco
	acknowledges the financial support of the European Research Council (grant SLING 819789), the AFOSR
	projects FA9550-18-1-7009, FA9550-17-1-0390 and BAA-AFRL-AFOSR-2016-0007 (European Office of
	Aerospace Research and Development), the EU H2020-MSCA-RISE project NoMADS - DLV-777826,
	and the Center for Brains, Minds and Machines (CBMM), funded by NSF STC award CCF-1231216.
	
	\printbibliography[title=References]
	\end{refsection}
	\newpage
	
	\begin{refsection}
	\appendix

	\section{Full Derivation of a Complexity Penalty for N-KRR}\label{app:derivation}
	
	\newcommand{\stilde}{\ensuremath{\widetilde{\Sigma}}}
	\newcommand{\sltilde}{\ensuremath{\widetilde{\Sigma}_{\lambda}}}
	\newcommand{\sltildeinv}{\ensuremath{\widetilde{\Sigma}_{\lambda}^{-1}}}
	\newcommand{\ghat}{\ensuremath{\widehat{g}_\lambda}}
	\newcommand{\gtilde}{\ensuremath{\widetilde{g}_\lambda}}
	\renewcommand{\S}{\Sigma}
	\newcommand{\Sl}{\Sigma_{\lambda}}
	\newcommand{\slinv}{\Sigma_{\lambda}^{-1}}
	\newcommand{\nysmap}{\widetilde{\Phi}_m}
	\newcommand{\kermap}{\Phi}
	\newcommand{\lemref}[1]{Lemma~\ref{#1}}
	
	We split the proof of Theorem~\ref{lem:main} into a few intermediate steps: after introducing the relevant notation and definitions we give a few ways in which the \nystrom{} estimator can be expressed,  useful in different parts of the proof. Then we proceed with three more technical lemmas, used later on. 
	We split the main proof into two parts to handle the two terms of the decomposition introduced in the main text of the paper: \lemref{lem:var1} for the sampling variance and \lemref{lem:var2} for the inducing point variance. Finally we restate Theorem~\ref{lem:main} for completeness, whose proof follows directly from the two variance bounds.

	\subsection{Definitions}
	
	Using the same notation as in the main text we are given data $\{(x_i, y_i)\}_{i=1}^n\subset\mathcal{X}\times\mathcal{Y}$ such that
	$$y_i = f^*(x_i) + \epsilon_i$$
	where $f^*: \mathcal{X}\rightarrow\mathcal{Y}$ is an unknown function, and the noise $\epsilon_i$ is such that $\expect{\epsilon_i} = 0, \expect{\epsilon_i^2} = \sigma^2$. We let $\hilbert$ be a RKHS and its subspace $\hm=\spa{\kf(z_1, \cdot), \dots, \kf(z_m, \cdot)}$ defined using the inducing points $\{z_j\}_{j=1}^m\subset\mathcal{X}$.
	We define a few useful operators, for vectors $\bm{v}\in\real{m}$ and $\bm{w}\in\real{n}$:
	\begin{align*}
		&\nysmap:\hilbert \rightarrow \real{m}, \quad \nysmap = (\kf(z_1, \cdot), \dots, \kf(z_m, \cdot)) \\
		&\nysmap^*: \real{m}\rightarrow\hilbert, \quad \nysmap^* \bm{v} = \sum_{j=1}^m\bm{v}_j\kf(z_j, \cdot) \\
		&\kermap:\hilbert \rightarrow \real{n}, \quad~~~ \kermap = (\kf(x_1, \cdot), \dots, \kf(x_n, \cdot)) \\
		&\kermap^*: \real{n}\rightarrow\hilbert, \quad~ \kermap^* \bm{w} = \sum_{i=1}^n\bm{w}_j\kf(x_i, \cdot).
	\end{align*}
	Let $\Sigma:\hilbert\rightarrow\hilbert = \kermap^*\kermap$ be the covariance operator, and $K=\kermap\kermap^*\in\real{n\times n}$ the kernel operator. 
	Further define $\knm = \kermap\nysmap^*\in\real{n\times m}$, $\kmm = \nysmap\nysmap^*\in\real{m\times m}$, and the approximate kernel $\kappr = \knm\kmm^\dagger\knm^\top\in\real{n\times n}$.
	The SVD of the linear operator $\nysmap$ is
	$$\nysmap = U\Lambda V^*$$
	with $U: \real{k}\rightarrow\real{m}$, $\Lambda$ the diagonal matrix of singular values sorted in non-decreasing order, $V:\real{k}\rightarrow\hilbert$, $k \le m$ such that $U^* U = I$, $V^* V = I$. 
	The projection operator with range $\hm$ is given by $P = VV^*$.

	The KRR estimator $\fh$ is defined as follows,
	$$\fh = \argmin_{f\in\hilbert}\frac{1}{n}\norm{f(X) - Y}^2 + \la\norm{f}_{\hilbert}^2.$$
	It can be shown~\citep{caponnetto07} that $\fh$ is unique and can be expressed in closed form as $\fh = \kermap^*(K + n\la I)^{-1}Y$.
	In the proofs, we will also use the noise-less KRR estimator, denoted by $\f$ and defined as,
	$$\f = \argmin_{f\in\hilbert}\frac{1}{n}\norm{f(X) - f^*(X)}^2 + \la \norm{f}_\hilbert^2.$$
	This estimator cannot be computed since we don't have access to $f^*$, but it is easy to see that
	$$\f = \kermap^*(K + n\la I)^{-1}f^*(X).$$	
	
	The \nkrr{} estimator, found by solving
	\begin{equation*}
		\nysfh = \argmin_{f\in\hm}\frac{1}{n}\norm{f(X) - Y}^2 + \lambda\norm{f}^2_{\hilbert}.
	\end{equation*}
	is unique, and takes the form (see \citet{lessismore}, Lemma 1)
	\begin{equation*}
		\nysfh = (P \S P + n\la I)^{-1} P \kermap^* Y
	\end{equation*}
	where $P$ is the projection operator with range $\hm$.
		
	The estimator $\nysfh$ can be characterized in different ways as described next.

	\subsection{Preliminary Results on the \nystrom{} estimator}
	The following lemma provides three different formulation of the \nystrom{} estimator. We will use the notation $A^\dagger$ to denote the Moore-Penrose pseudo-inverse of a matrix $A$.
	\begin{lemma}{(Alternative forms of the \nystrom{} estimator)}\label{lem:alt-nysfh}\\
		The following equalities hold
		\begin{align}
			\nysfh &= (P\S P + n\la I)^{-1} P \kermap^* Y \label{eq:nysf-form1}\\
			&= V(V^*\S V + n\la I)^{-1}V^*\kermap^* Y \label{eq:nysf-form2} \\
			&= \nysmap^*(\knm^\top \knm + \la n \kmm)^\dagger \knm^\top Y \label{eq:nysf-form3}
		\end{align}
		
		\textnormal{This Lemma is a restatement of results already found in the literature (e.g. in \citet{rudi2017falkon}, Lemmas 2 and 3) which are condensed here with slightly different proofs.}
		\begin{proof}
			Going from Eq.~\eqref{eq:nysf-form1} to Eq.~\eqref{eq:nysf-form2} consists in expanding $P = VV^*$ and applying the push-through identity
			\begin{align*}
				(P\S P + n\la I)^{-1}P\kermap^*Y  &= (VV^* \S VV^* + n\la I)^{-1}VV^* \kermap^*Y \\
				&= V(V^* \S VV^*V + n\la I)^{-1} V^* \kermap^*Y\\
				&= V(V^* \S V + n\la I)^{-1} V^* \kermap^*Y.
			\end{align*}
			
			To go from Eq.~\eqref{eq:nysf-form3} to Eq.~\eqref{eq:nysf-form2} we split the proof into two parts. We first expand Eq.~\eqref{eq:nysf-form3} rewriting the kernel matrices
			\begin{align*}
				\nysmap^*(\knm^\top \knm + \la n \kmm)^\dagger\knm^\top Y &= 
				\nysmap^*(\nysmap\kermap^*\kermap\nysmap^* +  n \la\nysmap\nysmap^*)^\dagger \knm^\top Y \\
				&= \nysmap^*(\nysmap(\S + n\la I)\nysmap^*)^\dagger\knm^\top Y.
			\end{align*}
			Then, we use some properties of the pseudo-inverse~\citep{pseudoinv01} to simplify $(\nysmap(\S + n\la I)\nysmap^*)^\dagger$,
			in particular, using the SVD of $\nysmap$, write
			\begin{equation*}
				(\underbrace{\vphantom{()}U\Lambda}_{F}
				\underbrace{V^*(\S+n\la I)V}_{H}
				\underbrace{\Lambda U^*\vphantom{()}}_{F^*})^\dagger.
			\end{equation*}
			Since $U$ has orthonormal columns, $F^\dagger = (U\Lambda)^\dagger =\Lambda^{-1} U^\dagger  = \Lambda^{-1} U^*$. A consequence is that $(F^*)^\dagger = (\Lambda U^*)^\dagger = (\Lambda^{-1}U^*)^* = U\Lambda^{-1}$.
			Then we split $(FHF^*)^\dagger$ into the pseudo-inverse of its three components in two steps.
			Firstly $(HF^*)^\dagger = (F^*)^\dagger H^\dagger$ if $H^\dagger H = I$ and $(F^*)(F^*)^\dagger = I$:
			\begin{enumerate}
				\item Since $H = V^*(\S + n\la I) V$ is invertible, $H^\dagger = H^{-1}$ and the first condition is verified.
				\item $F^* (F^*)^\dagger = \Lambda U^* U \Lambda^{-1} = I$.
			\end{enumerate}
			Also we have $(FHF^*)^\dagger = (HF^*)^\dagger F^\dagger $ if $F^\dagger F = I$ and $HF^*(HF^*)^\dagger = I$:
			\begin{enumerate}
				\item $F^\dagger F = \Lambda^{-1}U^*U\Lambda = I$,
 				\item $HF^*(HF^*)^\dagger = HF^*(F^*)^\dagger H^\dagger = HH^\dagger = I$.
			\end{enumerate}
			The end result of this reasoning is that
			\begin{equation*}
				(FHF^*)^\dagger = (F^*)^\dagger H^{-1} F^\dagger = U\Lambda^{-1}(V^*(\S + n\la I) V)^{-1}\Lambda^{-1}U^*
			\end{equation*}
			and hence
			\begin{align*}
				\nysmap^*(\knm^\top \knm + \la n \kmm)^\dagger\knm^\top Y &= V\Lambda U^*(U\Lambda V^*(\S + n\la I) V\Lambda U^*)^\dagger U\Lambda V^*\kermap^* Y \\
				&= V\Lambda U^* U\Lambda^{-1}(V^*(\S + n\la I) V)^{-1}\Lambda^{-1}U^* U\Lambda V^*\kermap^* Y \\
				&= V(V^*\S V + n\la I)^{-1} V^* \kermap^* Y
			\end{align*}
		\end{proof}
	\end{lemma}

	Another useful equivalent form, for the \nystrom{} estimator is given in the following lemma
	\begin{lemma}\label{lem:kequiv}
		Given the kernel matrices $\knm\in\real{n\times m}$, $\kmm\in\real{m\times m}$, and the \nystrom{} kernel $\kappr = \knm \kmm^\dagger \knm^\top\in\real{n\times n}$, the following holds
		\begin{equation}\label{eq:kequiv}
			(\kappr + n\la I)^{-1}\kappr = \knm(\knm^\top\knm + n\la\kmm)^\dagger\knm^\top
		\end{equation}
		\begin{proof}
			We state some facts about the kernel and image of the \nystrom{} feature maps
			\begin{equation*}
			\begin{gathered}
				(\ker\nysmap)^\perp = \spa{k(z_1, \cdot), \dots, k(z_m, \cdot)} = \Ima \nysmap^*  \\
				(\ker\nysmap^*)^\perp = \Ima \nysmap = \Ima \kmm = (\ker \kmm)^\perp = W\subseteq \real{m}.
			\end{gathered}
			\end{equation*}
			The space $\real{m}$ is hence composed of $\real{m} = W \oplus \ker \nysmap^*$. Take a vector $v\in\ker \nysmap^*$. We have that $\nysmap^* v = 0$, and $(\knm^\top\knm + n\la\kmm)v = \nysmap(\kermap^*\kermap + n\la I)\nysmap^* v = 0$.
			
			If instead $v\in W$, then $\nysmap(\kermap^*\kermap + n\la I)\nysmap^* v \in W$. Hence we have that
			$$\knm^\top \knm + n\la \kmm : W \rightarrow W$$
			and that $\kmm$ is invertible when restricted to the subspace $W$, but also $\knm^\top \knm + n\la\kmm$ is invertible on W. Furthermore by the properties of the pseudo-inverse, we have that
			\begin{equation}\label{eq:dag1}
				(\knm^\top\knm + n\la\kmm)(\knm^\top\knm + n\la\kmm)^\dagger = P_W
			\end{equation}
			with $P_W$ the projector onto set $W$.
			
			Furthermore we have the following equalities concerning the projection operator: $\kmm^\dagger\kmm = P_W$, as before; since $\knm = \kermap\nysmap^*$, $\knm P_W = \kermap\nysmap^* P_W = \knm$ and similarly its transpose
			$\knm^\top = \nysmap\kermap^*$ hence $P_W\knm^\top = \knm^\top$.
			
			Using these properties we can say
			\begin{align*}
				\knm\kmm^\dagger (\knm^\top\knm + n\la \kmm) &= \knm\kmm^\dagger\knm^\top\knm + n\la \knm\kmm^\dagger\kmm \\
					&= \knm\kmm^\dagger \knm^\top \knm P_W  + n\la \knm P_W \\
					&= (\knm\kmm^\dagger \knm^\top + n\la I) \knm P_W
			\end{align*}
			
			which implies that
			\begin{equation*}
				(\knm\kmm^\dagger\knm^\top + n\la I)^{-1} \knm\kmm^\dagger (\knm^\top\knm + n\la \kmm) = \knm P_W.
			\end{equation*}
			Multiplying both sides by $(\knm^\top\knm + n\la \kmm)^\dagger$, and using Eq.~\eqref{eq:dag1}
			\begin{equation}\label{eq:dag2}
				(\knm\kmm^\dagger\knm^\top + n\la I)^{-1}\knm\kmm^\dagger P_W = \knm P_W (\knm^\top\knm + n\la \kmm)^\dagger
			\end{equation}
			
			Hence we can write the left-hand side of our statement (Eq.~\eqref{eq:kequiv}), and use the properties of projection $P_W$ and Eq.~\eqref{eq:dag2} to get
			\begin{align*}
				(\knm\kmm^\dagger\knm^\top + n\la I)^{-1}\knm\kmm^\dagger\knm^\top &= (\knm\kmm^\dagger\knm^\top + n\la I)^{-1}\knm\kmm^\dagger P_W\knm^\top \\
				&= \knm P_W (\knm^\top\knm + n\la \kmm)^\dagger \knm^\top \\
				&= \knm (\knm^\top\knm + n\la \kmm)^\dagger \knm^\top
			\end{align*}
			which is exactly the right-hand side of our statement.
		\end{proof}
	\end{lemma}
	
	Finally, the algebraic transformation given in the following lemma allows to go from a form which frequently appears in proofs involving the \nystrom{} estimator ($\tr*{(I-P)\S}$) to a form which can easily be computed: the trace difference between the full and the \nystrom{} kernel.
	\begin{lemma}\label{lem:nystrace}
		Let $\nysmap: \hilbert\rightarrow\real{m}$ be the kernel feature-map of the inducing points with SVD $\nysmap = U\Lambda V^*$, such that the projection operator onto $\hm$ can be written $P = VV^*$. Also let $\kappr = \knm\kmm^\dagger\knm^\top$ be the \nystrom{} kernel. Then the following equivalence holds
		\begin{equation}
		\tr{(I - P)\S} = \tr{K - \kappr}.
		\end{equation}
		\begin{proof}
			Note that we can write $\kmm = \nysmap\nysmap^* = U\Lambda V^* V \Lambda U^* = U \Lambda^2 U^*$, which is a full-rank factorization since both $U\Lambda$ and $\Lambda U^\top$ are full-rank. Then we can use the formula for the full-rank factorization of the pseudoinverse (\citet{pseudoinv01}, Chapter 1, Theorem 5, Equation 24) to get
			\begin{align*}
			\kmm^{\dagger} &= (U\Lambda V^* V \Lambda U^*)^\dagger = (U\Lambda\Lambda U^*)^\dagger  \\
			&= U\Lambda (\Lambda U^* U \Lambda^2 U^* U \Lambda)^{-1}\Lambda U^* \\
			&= U\Lambda^{-2}U^*.
			\end{align*}
			
			Now we can prove the statement by expanding the left-hand side, and recalling $U^\top U = I$
			\begin{align*}
			\tr{(I - P)\S} &= \tr{(I - VV^*)\S} \\
			&= \tr{(I - V(\Lambda U^* U\Lambda^{-2} U^* U \Lambda)V^*)\kermap^* \kermap} \\
			&= \tr{\kermap(I - V\Lambda U^* (\nysmap\nysmap^*)^\dagger U\Lambda V^*)\kermap^*} \\
			&= \tr{\kermap\kermap^* - \kermap \nysmap^* (\nysmap\nysmap^*)^\dagger \nysmap\kermap^*} \\
			&= \tr{K - \knm \kmm^\dagger \knm^\top} = \tr{K - \kappr}.
			\end{align*}	
		\end{proof}
	\end{lemma}
	
	The following two lemmas provide some ancillary results which are used in the proof of the main lemmas below.
	\begin{lemma}\label{lem:projf}
		Let $P$ be the projection operator onto $\hm$, and $\f$ be the noise-less KRR estimator. Then the following bound holds
		\begin{equation}
			\norm{P\f}^2_{\hilbert} \le \norm{\f}^2.
		\end{equation}
		\begin{proof}
			This is a simple application of the definition of operator norm, coupled with the fact that orthogonal projection operators have eigenvalues which are either $0$ or $1$ (hence their norm is at most $1$).
			\begin{align*}
			\norm{P\f}^2_{\hilbert} &\le \norm{P}^2 \norm{\f}^2_{\hilbert} \\
			&\le \norm{\f}^2_{\hilbert}.
			\end{align*}
		\end{proof}
	\end{lemma}

	\begin{lemma}\label{lem:normf}
		Recall the notation $\errhatl{f} = n^{-1}\norm{f(X) - Y}^2 + \la\norm{f}_{\hilbert}^2$, and let $\f$ be the noise-less KRR estimator as before. Then the following statement holds:
		\begin{equation}
			\norm{\f}^2_\hilbert \le \expect{\dfrac{\errhatl{\f}}{\lambda}}
		\end{equation}
		where the expectation is taken with respect to the noise.
		\begin{proof}
			Recall that in the fixed design setting, given a fixed (i.e. not dependent on the label-noise) estimator, we always have
			$$\expect{\errhat{f}} = \err{f} + \sigma^2$$
			where $\sigma^2$ is the label-noise variance.
			
			In our case, noting that $L(\f)$ is always non-negative
			\begin{align*}
			\norm{\f}^2_\hilbert &= \dfrac{\la}{\la}\norm{\f}^2_\hilbert \\
			&\le \dfrac{\err{\f} + \la \norm{\f}^2_\hilbert}{\la} \\
			&\le \dfrac{\err{\f} + \sigma^2 + \la\norm{\f}^2_\hilbert}{\la} \\
			&= \dfrac{\expect{\errhatl{\f}}}{\la}.
			\end{align*}
		\end{proof}
	\end{lemma}
	
	\subsection{Proof of the main Theorem}
	The proof of Theorem~\ref{lem:main} starts from the error decomposition found in Section~\ref{sec:new-obj} which we report here:
	\begin{align*}
		& \expect{\err{\nysfh}} \le \mathbb{E}\Big[
		    \underbrace{\err{\nysfh} - \errhat{\nysfh}}_{\numcircledmod{1}} \\
		& + \underbrace{\errhat{\nysfh} + \reg{\nysfh} - \errhatl{P\f}}_{\numcircledmod{2}} 
		  + \underbrace{\errhatl{P\f}}_{\numcircledmod{3}} 
		\Big]
	\end{align*}
	and proceeds by bounding terms \numcircledmod{1} (see Lemma~\ref{lem:var1}) and \numcircledmod{3} (see Lemma~\ref{lem:var2}). After the two necessary lemmas we restate the proof of the main theorem which now becomes trivial.
	
	\begin{lemma}{(Bounding the data-sampling variance)}\label{lem:var1}\\
		Denoting by $\nysfh$ the \nkrr{} estimator, the expected difference between its empirical and test errors can be calculated exactly:
		\begin{equation*}
		\expect{\err{\nysfh} - \errhat{\nysfh}} = \dfrac{2\sigma^2}{n}\tr{(\kappr + n \la I)^{-1} \kappr}
		\end{equation*}
		with $\sigma^2$ the noise variance and $\kappr$ the \nystrom{} kernel.
		\begin{proof}
			For the sake of making the proof self-contained we repeat the reasoning of Section~\ref{sec:new-obj}.
			Starting with the expectation of the empirical error we decompose it into the expectation of the test error minus an inner product term:
			\begin{align*}
			\expect{\errhat{\nysfh}} &= \expect{\frac{1}{n}\norm{\nysfh(X) - f^*(X) - \epsilon}^2} \\
			&= \expect{\err{\nysfh}} + \sigma^2 - \dfrac{2}{n}\expect{\langle\nysfh(X) - f^*(X), \epsilon\rangle}.
			\end{align*}
			The $\sigma^2$ term is fixed for optimization purposes, so we must deal with the inner-product. We use the form of $\nysfh$ from Eq.~\eqref{eq:nysf-form3}, Lemma~\ref{lem:alt-nysfh}, and $\expect{\epsilon} = 0$, and to clean the notation we call
			$H \defeq \knm(\knm^\top\knm + n\la\kmm)^\dagger\knm^\top$:
			\begin{align*}
			\dfrac{2}{n}\expect{\langle\nysfh(X) - f^*(X), \epsilon\rangle} &= \dfrac{2}{n}\expect{\langle H(f^*(X) + \epsilon) - f^*(X), \epsilon\rangle} \\
			&= \dfrac{2}{n} \expect{\epsilon^\top H \epsilon} = \dfrac{2\sigma^2}{n}\tr{H},
			\end{align*}
			and using Lemma~\ref{lem:kequiv} $H$ can be expressed as $(\kappr+ n\la I)^{-1} \kappr$.

			Going back to the original statement we have
			\begin{align*}
				\expect{\err{\nysfh} - \errhat{\nysfh}} &= \expect{\err{\nysfh} - \err{\nysfh} + \dfrac{2\sigma^2}{n}\tr{(\kappr+ n\la I)^{-1} \kappr}} \\
					&= \dfrac{2\sigma^2}{n}\tr{(\kappr+ n\la I)^{-1} \kappr}
			\end{align*}
		\end{proof}
	\end{lemma}
	
	\begin{lemma}{(Bounding the \nystrom{} variance)}\label{lem:var2}\\
		Under the fixed-design assumptions, denote by $P$ the orthogonal projector onto space $\hm$, by $\errhatl{f}$ the regularized empirical risk of estimator $f$, and by $\f\in\hilbert$ the noise-less KRR estimator. Then the following upper-bound holds
		\begin{equation}
			\expect{\errhatl{P\f}} \le \dfrac{2}{n\la}\tr{K - \kappr}\expect{\errhatl{\f}} + 2\expect{\errhatl{\f}}.
		\end{equation}
		\begin{proof}
			Note that for estimators $f\in \hilbert$ we can always write $f(X) = \kermap f$. Hence for the projected KRR estimator
			we use that $(P\f)(X) = \kermap P\f$.
			We start by rewriting the left hand side to obtain a difference between projected and non-projected terms:
			\begin{align*}
			\expect{\errhat{P\f} + \reg{P\f}} &= 
				\expect{\frac{1}{n}\norm{\kermap P\f - f^*(X) - \epsilon}^2 + \reg{P\f}} \\
			&= \expect{\frac{1}{n}\norm{\kermap P\f - f^*(X)}^2 + \frac{1}{n}\norm{\epsilon}^2 + \reg{P\f}} \\
			&= \expect{\frac{1}{n}\norm{\kermap P\f - \kermap \f + \kermap \f - f^*(X)}^2 + \frac{1}{n}\norm{\epsilon}^2 + \reg{P\f}} \\
			&\le \expect{\frac{2}{n}\norm{\kermap P\f - \kermap \f}^2 + \frac{2}{n}\norm{\kermap \f - f^*(X)}^2 + \frac{2}{n}\norm{\epsilon}^2 + 2\reg{P\f}}
			\end{align*}
			where we used the fact that $\expect{\epsilon} = 0$, and the triangle inequality in the last step.
			
			By Lemma~\ref{lem:projf}, and the definition of $\expect{\errhat{f}}$ we have that
			\begin{equation*}
			\expect{\frac{2}{n}\norm{\kermap \f - f^*(X)}^2 + \frac{2}{n}\norm{\epsilon}^2 + 2\reg{P\f}} \le 2\expect{\errhat{\f}}.
			\end{equation*}
			
			Next we use again the definition of operator norm to deal with the difference between projected and non-projected noise-less KRR estimators:
			\begin{align*}
			\expect{\frac{2}{n} \norm{\kermap P\f - \kermap \f}^2} &= 
				\frac{2}{n}\norm{\kermap(P - I)\f}^2 \\
			&\le \frac{2}{n}\norm{\kermap(I - P)}^2\norm{\f}^2.
			\end{align*}
			The first part of this latter term is
			\begin{equation*}
			\norm{\kermap(I-P)}^2 = \norm{(I - P)\kermap^\top\kermap (I - P)} \le \tr{(I - P)\kermap^\top\kermap} = \tr{(I-P)\S}
			\end{equation*}
			since the trace norm controls the operator norm, and using the cyclic property of the trace and the idempotence of the projection operator $I-P$. By Lemma~\ref{lem:nystrace} we have that $\norm{\kermap(I-P)}^2 \le \tr{K-\kappr}$.
			For the second part we use Lemma~\ref{lem:normf} so that
			\begin{align*}
			\norm{\f}^2 \le \expect{\dfrac{\errhatl{\f}}{\lambda}}
			\end{align*}
			which concludes the proof.
		\end{proof}
	\end{lemma}

	We now have all the ingredients to prove Theorem~\ref{lem:main} which we restate below for the reader.
	\begin{theorem*}{(Restated from the main text)}\\
		Under the assumptions of fixed-design regression we have that,
		\begin{align}
		\expect{\err{\nysfh}} \le &\nonumber \dfrac{2\sigma^2}{n}\tr{(\kappr + \la I)^{-1} \kappr} \nonumber \\
		& + \dfrac{2}{n\la}\tr{K - \kappr} \expect{\errhat{\f}} \nonumber \\
		& + 2\expect{\errhat{\f}} \label{eq:ub-app}
		\end{align}
		\begin{proof}
			The decomposition is the same:
			\begin{align*}
				& \expect{\err{\nysfh}} \le \mathbb{E}\Big[\underbrace{\err{\nysfh} - \errhat{\nysfh}}_{\numcircledmod{1}} \\
				& + \underbrace{\errhat{\nysfh} + \reg{\nysfh} - \errhatl{P\f}}_{\numcircledmod{2}} 
				+ \underbrace{\errhatl{P\f}}_{\numcircledmod{3}} \Big]
			\end{align*}
			where $\numcircledmod{2} \le 0$. We may then use Lemma~\ref{lem:var1} for term \numcircledmod{1} and Lemma~\ref{lem:var2} for term \numcircledmod{3} to obtain
			\begin{align*}
				& \expect{\err{\nysfh}} \le 
					\dfrac{2\sigma^2}{n}\tr{(\kappr + n\la I)^{-1}\kappr} + 
					\dfrac{2}{n\la}\tr{K-\kappr}\expect{\errhatl{\f}} + 2\expect{\errhatl{\f}}.
			\end{align*}
		\end{proof}
	\end{theorem*}

\section{Datasets}\label{app:datasets}

We used a range of datasets which represent a wide spectrum of scenarios for which kernel learning can be used. They can be divided into three groups: medium sized unstructured datasets (both for regression and binary classification), medium sized image recognition datasets (multiclass classification) and large unstructured datasets (classification and regression).
We applied similar preprocessing steps to all datasets (namely standardization of the design matrix, standardization of the labels for regression datasets, one-hot encoding of the labels for multiclass datasets). When an agreed-upon test-set existed we used it (e.g. for MNIST), otherwise we used random 70/30 or 80/20 train/test set splits, with each experiment repetition using a different split. Below we provide more details about the datasets used, grouping several of them together if the same procedures apply. 
The canonical URLs at which the datasets are available, along with their detailed dimensions and training/test splits are shown in Table~\ref{tab:dsets}

The error metrics used are dataset-dependent, and outlined below. For regression problems we use the RMSE, defined as $\sqrt{n^{-1} \sum_{i=1}^n (y_i - \hat{f}(x_i))^2}$ and its normalized version the NRMSE:
$$NRMSE : \autovbar*{\dfrac{\sqrt{\frac{1}{n}\sum_{i=1}^n (y_i - \hat{f}(x_i))^2}}{\frac{1}{n}\sum_{i=1}^n y_i}}.$$
For classification problems we use the fraction of misclassified examples (c-error), and the area under the curve (AUC) metric.

\paragraph{SpaceGA, Abalone, MG, CpuSmall, Energy }
Small regression datasets between 1385 (MG) and 8192 (CpuSmall) samples, label standardization is performed and error is measured as NRMSE. The predictor matrix is also standardized.

\paragraph{Road3D, Buzz, Protein, HouseElectric, BlogFeedback}
Regression datasets of medium to large size from the UCI ML repositories. We used label standardization for Road3D, BlogFeedback, Buzz and Protein, and an additional log transformation for HouseElectric. Measured error is NRMSE. The predictor matrix is standardized.

\paragraph{MNIST, FashionMNIST, SVHN, CIFAR-10}
Four standard image recognition datasets. Here the labels are one-hot encoded (all datasets have 10 classes), and the design matrix is normalized in the 0-1 range. Standard train/test splits are used.

\paragraph{Chiet}
A time-series dataset for short-term wind prediction. The labels and predictors are standardized, and the error is measured with the NRMSE. A fixed split in time is used.

\paragraph{Ictus}
A dataset simulating brain MRI. Predictors are standardized and a random 80/20 split is used.

\paragraph{Cod-RNA, SVMGuide1, IJCNN1, CovType}
Four datasets for binary classification ranging between approximately \num{3000} points for SvmGuide1 and \num{5e5} points for CovType. The design matrix is standardized while the labels are $-1$ and $+1$.

\paragraph{Higgs, SmallHiggs}
HIGGS is a very large binary classification dataset from high energy physics. We took a small random subsample to generate the SmallHiggs dataset, which has predefined training and test sets. The design matrix is normalized by the features' variance.
For the HIGGS dataset we measure the error as $1$ minus the AUC.

\paragraph{Flights, Flights-Cls}
A regression dataset found in the literature~\citep{hensman13, vffgp_hensman17} which can also be used for binary classification by thresholding the target at 0.

\begin{table}[ht!]
	\caption{Key details on the datasets used.} \label{tab:dsets}
	\centering
	\begin{tabular}{lllll}
		\toprule
		& n & d & train/test & error  \\
		\midrule
		
		\href{https://www.csie.ntu.edu.tw/~cjlin/libsvmtools/datasets/regression.html}{SpaceGA} &
		\num{3107} & \num{6} & 70\%/30\% & NRMSE \\
		
		\href{https://www.csie.ntu.edu.tw/~cjlin/libsvmtools/datasets/regression.html}{Abalone} &
		\num{4177} & \num{8} & 70\%/30\% & NRMSE \\
		
		\href{https://www.csie.ntu.edu.tw/~cjlin/libsvmtools/datasets/regression.html}{MG} &
		\num{1385} & \num{6} & 70\%/30\% & NRMSE \\
		
		\href{https://www.csie.ntu.edu.tw/~cjlin/libsvmtools/datasets/regression.html}{CpuSmall} &
		\num{8192} & \num{12} & 70\%/30\% & NRMSE \\
		
		\href{https://archive.ics.uci.edu/ml/machine-learning-databases/00374/}{Energy} &
		\num{768} & \num{8} & 80\%/20\% & NRMSE \\
		
		\href{https://archive.ics.uci.edu/ml/datasets/3D+Road+Network+(North+Jutland,+Denmark)}{Road3D} &
		\num{434874} & \num{3} & 70\%/30\% & RMSE \\
		
		\href{https://archive.ics.uci.edu/ml/datasets/Buzz+in+social+media}{Buzz} &
		\num{2049280} & \num{11} & 70\%/30\% & RMSE \\
		
		\href{https://archive.ics.uci.edu/ml/datasets/Physicochemical+Properties+of+Protein+Tertiary+Structure}{Protein} &
		\num{45730} & \num{9} & 80\%/20\% & NRMSE \\
		
		\href{https://archive.ics.uci.edu/ml/machine-learning-databases/00304/}{BlogFeedback} &
		\num{60021} & \num{280} & \num{52397}/\num{7624} & RMSE \\
		
		\href{https://yann.lecun.com/exdb/mnist/}{MNIST} &
		\num{70000} & \num{784} & \num{60000}/\num{10000} & 10 class c-error \\
		
		\href{https://github.com/zalandoresearch/fashion-mnist}{FashionMNIST} &
		\num{70000} & \num{784} & \num{60000}/\num{10000} & 10 class c-error \\
		
		\href{http://ufldl.stanford.edu/housenumbers/}{SVHN} &
		\num{99289} & \num{1024} & \num{73257}/\num{26032} & 10 class c-error \\
		
		\href{https://www.cs.toronto.edu/~kriz/cifar.html}{CIFAR-10} &
		\num{60000} & \num{1024} & \num{50000}/\num{10000} & 10 class c-error \\
		
		Chiet & \num{34059} & \num{144} & \num{26227}/\num{7832} & NRMSE \\
		
		Ictus & \num{29545} & \num{992} & 80\%/20\% & binary c-error \\
		
		\href{https://www.csie.ntu.edu.tw/~cjlin/libsvmtools/datasets/binary.html}{Cod-RNA} &
		\num{331152} & \num{8} & \num{59535}/\num{271617} & binary c-error \\
		
		\href{https://www.csie.ntu.edu.tw/~cjlin/libsvmtools/datasets/binary.html}{SVMGuide1} &
		\num{7089} & \num{4} & \num{3089}/\num{4000} & binary c-error \\
		
		\href{https://www.csie.ntu.edu.tw/~cjlin/libsvmtools/datasets/binary.html}{IJCNN1} &
		\num{141691} & \num{22} & \num{49990}/\num{91701} & binary c-error \\
		
		\href{https://www.csie.ntu.edu.tw/~cjlin/libsvmtools/datasets/binary.html}{CovType} &
		\num{581012} & \num{54} & 70\%/30\% & binary c-error \\
		
		\href{https://archive.ics.uci.edu/ml/datasets/HIGGS}{SmallHiggs} &
		\num{30000} & \num{28} & \num{10000}/\num{20000} & binary c-error \\
		
		\href{https://archive.ics.uci.edu/ml/datasets/HIGGS}{Higgs} &
		\num{1.1e7} & \num{20} & 80\%/20\% & 1 - AUC \\
		
		\href{https://www.transtats.bts.gov/Fields.asp?Table_ID=236}{Flights} &
		\num{5.93e6} & \num{8} & 66\%/34\% & MSE \\
		
		\href{https://www.transtats.bts.gov/Fields.asp?Table_ID=236}{Flights-Cls} &
		\num{5.93e6} & \num{8} & \num{5829413}/\num{100000} & binary c-error \\
		\bottomrule
	\end{tabular}
\end{table}

\section{Experiment Details}\label{app:experimental}

All experiments were run on a machine with a single NVIDIA Quadro RTX 6000 GPU, and 256GB of RAM. The details of all hyperparameters and settings required to reproduce our experiments are provided below. Relevant code is available in the repository at \url{https://github.com/falkonml/falkon}.

\subsection{Small scale experiments}
We ran the small scale experiments by optimizing the exact formulas for all objectives, computed with Cholesky decompositions and solutions to triangular systems of equations.
We used the Adam optimizer with default settings and ran it for \num{200} epochs with a fixed learning rate of \num{0.05}. We optimized $m=100$ inducing points initialized to the a random data subset, used the Gaussian kernel with a separate length-scale for each data-dimension (the initialization using the median heuristic was the same for each dimension), and the amount of regularization $\lambda$ which was initialized to $1/n$.
The validation set size (for the \textit{Hold-out} objective) was fixed to 60\% of the full training data. While this may seem large, in our setting the size of the hyperparameter space (in first approximation $m\times d$) is larger than the number of model parameters ($m\times o$ where $o$ is the dimension of the target space $\mathcal{Y}$, most commonly $o=1$).

\subsection{Large scale experiments}
We ran the large-scale experiments for just the $\mathcal{L}^{\mathrm{Prop}}$ objective, while the other performance numbers in Table~\ref{tab:large-scale} are taken from~\citet{billions}.
For our objective we again used the Adam optimizer. For the Flights and Higgs dataset we trained with learning rate \num{0.05} for \num{20} epochs, while we trained Flights-Cls with a smaller learning rate of \num{0.02} for \num{10} epochs.
We used the Gaussian kernel with a single length-scale, initialized as in~\citep{billions} (Flights $\sigma_0=1$, Flights-Cls $\sigma_0=1$, Higgs $\sigma_0=4$) and $\lambda_0 = 1/n$. We used $t=20$ stochastic trace estimation vectors for all three experiments, sampling them from the standard Gaussian distribution. The STE vectors were kept fixed throughout optimization.
The conjugate gradient tolerance for the Falkon solver was set to \num{5e-4} for Flights-Cls, and \num{1e-3} for Flights and Higgs (a higher tolerance corresponds to longer training time), while we always capped the number of Falkon iterations to \num{100}.

	\printbibliography[title=References]
	\end{refsection}

\end{document}